\documentclass{article}

\usepackage{amsmath}
\usepackage{amssymb}
\usepackage{float}
\usepackage{scalefnt}
\usepackage{ifthen}
\usepackage{url}
\usepackage{graphicx}
\usepackage{colortbl}
\usepackage{multirow}
\usepackage{array}
\usepackage{listings}
\newtheorem{myproposition}{myproposition}

\newcommand{\PR}{{\tt P}}
\newcommand{\PRl}{{\tt PartOf}}

\newcommand{\CR}{{\tt C}}
\newcommand{\CRl}{{\tt Connected}}

\newcommand{\OR}{{\tt O}}
\newcommand{\ORl}{{\tt Overlap}}

\newcommand{\Un}{\mathcal{U}}

\begin{document}

%\catchline{}{}{}{}{}
%
%%%%%%%%%%%%%%%%%%%%%%%%%%%%%%%%%%%%%%%%%%%%%%%%%%%%%%%%%%%%%%%%%%%%

\title{Symbol Grounding via Chaining of Morphisms}

\author{ Ruiting Lian$^{1,2,3}$ and  Ben Goertzel$^{1,3}$  and  Linas Vepstas$^{1,3}$  \\
and David Hanson$^{2}$ and Changle Zhou$^{1,4}$  \\ 
$^1$ Dept. of Cognitive Science, Xiamen University  \\ $^2$  
Hanson Robotics $^3$ OpenCog Foundation $^4$ {\it corresponding author}}

\maketitle

\begin{abstract}
A new model of symbol grounding is presented, in which the structures of natural language, logical semantics, perception and action are represented categorically, and symbol grounding is modeled via the composition of morphisms between the relevant categories.    This model gives conceptual insight into the fundamentally  systematic nature of symbol grounding, and also connects naturally to practical real-world AI systems in current research and commercial use.  Specifically, it is argued that the structure of linguistic syntax can be modeled as a certain asymmetric monoidal category, as e.g. implicit in the link grammar formalism; the structure of spatiotemporal relationships and action plans can be modeled similarly using "image grammars" and "action grammars"; and common-sense logical semantic structure can be modeled using dependently-typed lambda calculus with uncertain truth values.  Given these formalisms, the grounding of linguistic descriptions in spatiotemporal perceptions and coordinated actions consists of following  morphisms from language to logic through to spacetime and body (for comprehension), and vice versa (for generation).   The mapping is indicated between the spatial relationships in the Region Connection Calculus and Allen Interval Algebra and corresponding entries in the link grammar syntax parsing dictionary.   Further, the abstractions introduced here are shown to naturally model the structures and systems currently being deployed in the context of using the OpenCog cognitive architecture to control Hanson Robotics humanoid robots.   
\end{abstract}

%\keywords{symbol grounding, category theory, morphism, compositionality, semiotics, image grammar, RCC, Allen interval algebra} 

%\Papertype{\textbf{Research paper}}

\section{Introduction}
The concept of "symbol grounding"  -- the formation and manipulation of correspondences between linguistic tokens used by an agent, and perceptions and actions in the agent's physical environment -- was originally formulated by Searle \cite{Searle1980} and Newell \cite{Newell1990} nearly four decades ago, and was first crisply summarized and brought to the scientific community's broad attention by Harnad in 1990 \cite{harnad1990}.   By now the notion is no longer so controversial, and it is fairly widely accepted that symbol grounding is, in some sense, a critical aspect of creating AI systems that use natural language effectively and with full understanding \cite{Cangelosi2011}.    The concept is especially firmly embraced in the realm of cognitive robotics \cite{Coradeschi2013}, where one confronts the practical problem of mapping the linguistic tokens implicit in the robot's speech output and input, with the continuous-variable patterns of vision, audition, haptics, kinesthetics and movement that the robot experiences.   For cognitive roboticists experimenting with linguistically interacting robots, symbol grounding is not just a puzzle of philosophy and cognitive science, but an everyday hands-on R\&D struggle.

In neither cognitive science nor cognitive robotics, however, is there currently a clear consensus on what "symbol grounding" actually means, when one gets beyond the general concept and digs into the details.   Clearly, symbol grounding is more than just mapping discrete words onto classes of objects or events -- e.g. the word "apple" onto observations of apples, or the word "walk" onto observations of entities walking.    As Cangelosi has pointed out \cite{Cangelosi2011}, the next frontier in understanding symbol grounding appears to be coming to grips with its {\it systematic} nature, with the way in which networks of linguistic relationships are grounded by networks of relationships in the non-linguistic world.   A recent European government funded research initiative pursues precisely this topic \footnote{\url{http://www.chistera.eu/projects/reground}}, with a team combining expertise in robotics, cognitive science and computational linguistics, and a focus on grounding appropriate linguistic expressions in affordances pertaining to actions in simple simulation environments.

Here we model the systematic nature of symbol grounding in a novel way, using the framework of category theory to represent the abstract structures implicit in logic, language, sensory data (visual in particular) and motor actions, and to explore the sorts of mappings that exist between these different domains.   Our closest predecessor in this investigation is Goguen \cite{goguen2005}, who noted the relevance of category-theoretic notions for modeling the mappings between linguistic domains and concrete grounding domains.   However, this interesting prior work remained at a very theoretical level, whereas our goal here is to use these abstract notions to model our current practical work on computational language processing, sensory data processing and robot movement control.   

The systematic nature of symbol grounding is something we face quite concretely in our work using the OpenCog cognitive architecture to control Hanson Robotics humanoid robots (focusing mainly on face, head and neck control at the moment, though we have also worked with robots with torsos and complete walking bodies) \cite{Goertzel2014c}.  Making spoken natural language dialogue, machine vision and audition and coordinated physical action work effectively in a real-world social robotics context involves a great variety of complex details, and it is valuable to have a clear overarching framework in which to position all the details.

Building on prior work by ourselves and others in multiple relevant domains, we argue here that:

\begin{itemize}
\item the crux of syntactic structure, as modeled in link grammar or pregroup grammar, is captured by modeling syntax as certain asymmetric monoidal categories
\item logical semantics is adequately captured via dependently typed lambda calculus with two-valued truth values, which then corresponds to a certain locally closed Cartesian category
\item perceptual structure may be captured by "image grammars" and their generalizations, which can be modeled via the same categories used to model linguistic syntax, with the role of "parts of speech" played by "types of object or identified object-part."   For instance, these perceptual grammars can be formed, in practice, via pattern mining on the states of deep neural networks.
\item the structure of movement-control actions may be modeled similarly to perceptions,with the role of "parts of speech" played by "animations" -- coherent movement-patterns involving one or more actuators
\end{itemize}

\noindent Given these formalizations, we show that symbol grounding can be effectively modeled via morphisms from natural language syntax to logic, and then from logic to perception or action.

A key point is that we are modeling symbol grounding not merely as a collection of mappings from individual words or linguistic relationships onto individual classes of percepts or actions, or individual relations therebetween.  Rather, we are modeling symbol grounding as a systematic mapping from the algebra of language into the algebra of logic, and then from the algebra of logic into the algebras of perception and action.   By composing these morphic mappings, one grounds language in perception and actions, and generally maps perception, action, language and logic into each other.   

It also is possible, using these tools, to model grounding of language directly in perception and action, without need for the intervening medium of logic.   We suggest that this sort of grounding does in fact occur in some cases.  However, we suggest that in the majority of cases, if an agent needs to carry out complex activities, the (explicit or implicit) use of logical algebra as a connecting medium between perception, action and language will be the most effective strategy in practice.   This is in part because logical algebra contains powerful means for abstraction and generalization, so that including the power of logic in the morphism pipeline provides a clear route for a system to extend its already-learned perceptual, movement and language procedures to new and different situations.

From a theoretical linguistics view, the perspective presented here represents a sort of blend of the Saussurean view of linguistic meaning as comprised of language-internal relationships \cite{Saussure2013}, and the view of linguistic meaning as comprised of mappings between linguistic entities and external entities such as physical or social phenomena \cite{tomasello2003}.   Linguistic meaning is proposed to have to do with mappings between the system of language-internal relationships, and the system of relationships between entities in the world.   At a broad level there is resonance between this perspective and, for instance, systemic-functional grammar \cite{Halliday2004, Mwinlaaru2016}, but the particulars of the approach we pursue here are quite different.

%%%%%%%%%%%%%%%%%%%%%%%%%%%%%%%%%%%%%%%%%%%%%%%%%%%%%%%%%%%%%%%%%%%%%%%%%%%%

\section{A Categorial View of Natural Language Comprehension and Generation}

The first step toward articulation of the mappings presented here, is the interpretation of natural language syntax as categorial in nature.   Such an understanding has been provided by the work of Lambek and colleagues \cite{Lambek2008, Lambek2006}, who have outlined "pregroup grammars" that appear to explain phenomena of English syntax and that are also equivalent to asymmetric monoidal categories.   

There have not yet been any practical computational linguistic systems based explicitly on pregroup grammars.   However, Linas Vepstas \footnote{\url{http://www.abisource.com/projects/link-grammar/dict/introduction.html}} \footnote{\url{https://en.wikipedia.org/wiki/Link_grammar}}has made the observation that there is an elementary equivalence between pregroup grammars and the link grammar, a grammatical framework conceived by Sleator and Temperley \cite{Sleator1993}, which comes along with a rather thorough hand-coded lexicon for English, and useful though less complete lexicons for several other languages.  There is current research work on replacing these hand-coded lexicons with comparable lexicons based on unsupervised learning (if successful, this will yield lexicons with even broader coverage) \cite{Vepstas2014}.

\subsection{From Link Grammar to Pregroup Grammar}

Link grammar is closely related to standard dependency grammars, but also has some significant differences.   The essential idea of link grammar is that each word comes with a feature structure consisting of a set of typed connectors.  Parsing consists of matching up connectors from one word with connectors from another.  

For simplicity we will restrict attention to English in our discussion here, but the link grammar also covers a broad variety of languages.  In particular, while the examples given here involve connectors at the word level, it can also be extended to languages with concatenative morphology, via positing connections at the morpheme level (see the link grammar dictionaries for e.g. Hebrew and Turkish for examples).

Consider the following example, drawn from the classic paper ``Parsing with a Link Grammar''  by Sleator and Temperley \cite{Sleator1993}:

\begin{center}
\textbf{The cat chased a snake.}
\end{center}

\noindent The link grammar parse structure for this sentence is:

\includegraphics[width=12cm]{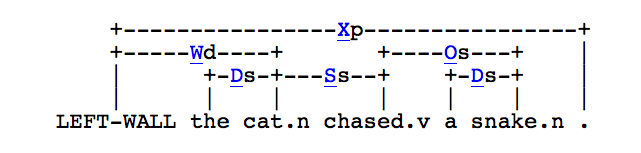}

\noindent It turns out to be useful to imagine that there was a dummy word at the beginning of every sentence, denoted "LEFT-WALL"; this is used, e.g., to identify the head of a sentence.  There is also a corresponding "RIGHT-WALL," which is used only for certain punctuation phenomena. In most sentences, we use a special "RW" connector to simply connect punctuation and the "RIGHT-WALL". 

The ``link grammar dictionary''  contains connectors associated with all common English words.  The notation used to describe feature structures in this dictionary is quite simple.  Different kinds of connectors are denoted by letters or pairs of letters like S or SX.  Then if a word W1 has the connector S+, this means that the word can have an S link coming out to the right side.  If a word W2 has the connector S-, this means that the word can have an S link coming out to the left side.  In this case, if W1 occurs to the left of W2 in a sentence, then the two words can be joined together with an S link.  

The features of the words in our example sentence, as given in the classic S\&T paper, are

\begin{tabular}{|p{2.0in}|p{2.1in}|} \hline 
\textbf{Words} & \textbf{Formula} \\ \hline 
a, the & D+ \\ \hline 
snake, cat & D- \& (O- or S+) \\ \hline 
Chased & S- \& O+ \\ \hline 
\end{tabular}

\noindent The nature of linkage imposes constraints on the variable assignments; for instance, if ``the''  is assigned as the value of the ``word that links to D-''  feature of ``snake'' , then ``snake''  must be assigned as the value of the ``word that links to D+''  feature of ``the.''    

The rules of link grammar impose additional constraints -- i.e. the planarity, connectivity, ordering and exclusion metarules described in Sleator and Temperley's papers.  In essence, what these constraints mean is that a link parse consists of a set of arcs drawn over a sentence, where each arc leads from one word to another, the set of words and arcs forms a connected graph, and no two arcs cross each other when drawn in the plane.  

There may be unusual cases where it is desirable to allow links to cross, and to encompass such cases one can introduce a broader constraint called "landmark transitivity," as explored in Hudson's "word grammar" framework \cite{Hudson2007}.  However, this is still linguistically a bit controversial, and so we will ignore this from the standpoint of this paper, and stick with the "no links cross" constraint.  Landmark transitivity can be handled via a slight extension of the formalism presented here, which is left as an exercise for the reader.

To see how link grammar is equivalent to a categorial perspective on grammar, consider that each word can be viewed as a transformation that maps (usually other) words into phrases.    So for instance if we have

{\tt\begin{small}\begin{verbatim}
   +-- S+ --
   |
  dogs
\end{verbatim}\end{small}}

\noindent i.e. "dogs" with an S connector on the right -- this is a transformation
that transforms {any word with an S connector on its left} into a
phrase.   We may have

{\tt\begin{small}\begin{verbatim}
   -- S- --+
           |
          bark
\end{verbatim}\end{small}}

\noindent in which case if we let "dogs" transform "bark", it will transform it into

{\tt\begin{small}\begin{verbatim}
   +-- S ---+
   |        |
dogs       bark
\end{verbatim}\end{small}}

\noindent i.e. into the phrase "dogs bark."

From here it is a simple step to consider the process of
connecting a word to another word or phrase as a {\it product}, in the categorial
sense.  Then
the entity "dog" as an entity to connect to the left of another word or
phrase is the right adjoint of "dog"; and the entity "dog" as an
entity to connect to the right of another word, is the left adjoint of
"dog", and so we have a monoidal category.  Which is evidently an
asymmetric monoidal category -- because grammar is left-right
asymmetric: a word's potential syntactic connections on the left are
not the same as its potential syntactic connections on the right.  
That is, the left and right going connectors in the link grammar dictionary 
correspond to left and right adjuncts in the categorial representation; so that the asymmetry at the categorial level
simply reflects the left-right nature of written language, which reflects the 
embedding of language in one-dimensional time

To model practical parsing using this framework, one can introduce the notion of a "constrained transformation
system" -- where we have a certain set of objects, and then a certain
set of transformations, and a set of constraints on which
transformations can be used together.   The objects plus
transformations can be thought of as a category.   We then have a
set of disjunctions among (object, transformation) pairs, indicating
that if e.g. $(O1,T1)$ is in the transformation system, then $(O2,T2)$
cannot be.   These disjunctions serve as constraints.

For instance, the disjunctive constraints in the link grammar dictionary for
"bark" would contain the restriction that if we apply the
word-instance "dogs" to the word-instance "bark" as indicated above,
we can't also apply the same word-instance "dogs" to the word-instance "me"
 (as in the case "This sentence dogs me" ).   

\subsection{A Categorial View of Logical Semantics for Natural Language}

Semantics, like syntax, can also be modeled categorically.  There are many
different semantic frameworks to choose from; for the sake of discussion here,
we will choose the framework that is currently in practical use in the OpenCog AI framework, operating
together with the link grammar formalism.   This framework is called Probabilistic Logic
Networks \cite{Goertzel2008}, and from a mathematical perspective it can be viewed as predicate logic augmented
with a labeling of terms and predicates with uncertain truth value objects representing imprecise probabilities.  In the simplest case
the truth value objects consist of $(s,n)$ pairs, where $s$ connotes a probability and $n$ connotes a nonnegative "weight of evidence".  These pairs are imprecise probabilities a la Walley \cite{Walley1991}; extended versions involve indefinite or distributional probabilities \cite{Ikle2007}, but these extensions don't change the basic categorical picture presented here so can be ignored for the present.

As articulated in \cite{Goertzel2008}, PLN is not fully formalized but only semi-formalized.
From a formal perspective,  PLN can be viewed
as building on typed lambda calculus, adding onto it imprecise-probabilistic
truth values attached to expressions.   In its full generality PLN
should be viewed as building on lambda calculus with {\it dependent} types.

Just as simply typed lambda calculus corresponds to closed Cartesian
categories \footnote{ \url{http://www.goodmath.org/blog/2012/03/11/interpreting-lambda-calculus-using-closed-cartesian-categories/}}; similarly, lambda calculus with dependent types is known to
correspond to a (locally closed) cartesian category.   \footnote{ A category C is
locally Cartesian closed, if all of its slices C/X are Cartesian
closed (i.e. if it's Cartesian closed for each parameter value used in
a parameter-dependent type)}.  An arrow $F: X\rightarrow Y$  can be interpreted as
a variable substitution and as a family of types indexed over Y in the
type theory.   Basically, the category associated with a typed lambda
calculus has objects equal to the types, and morphisms equal to
type-inheritance relations between the types.

 Note that unlike the asymmetric monoidal category used to model
syntax, here we have a {\it symmetric} category modeling logic expressions.
This is, crudely, because logic doesn't care about left versus right,
whereas link grammar (and syntax in general) does.  Logic cares about
predicate versus argument; but syntax has forms in which predicate comes after
argument and forms in which predicate comes before argument, whereas logic
does not contain this distinction.

\subsection{Mapping Language to Logic}

Language and logic, according to the formalisms summarized above, are formal
systems with different, but closely related and in some senses parallel structure.
One way to model the interpretation of language is in terms of formal mappings
into logical formal structures from linguistic formal structures, and corresponding
mappings into logical formal structures from formal structures characterizing
nonlinguistic domains to which language refers.

There exist multiple formalisms described in the research
 literature, and implemented in software, for mapping linguistic structures such as
 syntactic parse trees into logical formalisms.   Two relatively well fleshed out examples
 are Fluid Construction Grammars \cite{Steels2006} and the language processing front end of the
 (closed-source and proprietary) Cyc AI system \cite{Lenat1990}.   Here we will describe the syntax-to-logical-semantics
 mapping system we have developed in the context of the OpenCog software system, which 
 transforms link grammar parses of sentences into sets of logic expressions consistent with the
 Probabilistic Logic Networks framework.
 
In the OpenCog software system, this mapping from syntax to logic is contained in a series of two
rule systems called RelEx and RelEx2Logic. 
RelEx and RelEx2Logic, as a pipeline, maps syntactic structures into logical structures.    Currently these rule systems comprise
hand-coded rules; work is underway aimed at replacing these with rules obtained
via supervised learning on parallel corpora involving the constructed logical language
Lojban \cite{Cowan2016} and natural languages.

\subsubsection{RelEx}

RelEx \footnote{RelEx was originally designed and implemented by Mike Ross and Ben Goertzel, but has since been heavily modified by others, including Linas Vepstas, Ruiting Lian, Ben Goertzel and Alex van der Peet}.takes as input the link parses of sentences, and produces as output simplified representations of the grammatical structure of input sentences. Some surface details of grammatical structure are normalized.   

For example, the current version of RelEx, applied to the standard example "The cat chased a snake." given above, produces the output:

{\tt\begin{small}\begin{verbatim}
 Dependency relations:
 
     _obj(chase, snake)
     _subj(chase, cat)

 Attributes:

     tense(chase, past)
     subscript-TAG(chase, .v-d)
     pos(chase, verb)
     pos(., punctuation)
     subscript-TAG(snake, .n)
     pos(snake, noun)
     noun_number(snake, singular)
     definite-FLAG(cat, T)
     subscript-TAG(cat, .n)
     pos(cat, noun)
     noun_number(cat, singular)
     pos(a, det)
     pos(the, det)
\end{verbatim}\end{small}}

Via the action of RelEx, many equivalent verb-frame argument assignments are mapped to identical or graphically homomorphic representations. This provides a certain degree of semantic normalization.  For example,  comparing

{\tt\begin{small}\begin{verbatim}
The cat chased a snake.
A snake was chased by a cat.
\end{verbatim}\end{small}}

\noindent both sentences share the following RelEx dependencies:

{\tt\begin{small}\begin{verbatim}
     _subj(chase, snake)
     _obj(chase, cat)
\end{verbatim}\end{small}}

Broadly speaking the output of RelEx somewhat resembles that of the Stanford parser \cite{marneffe2006}; and there is a Stanford-format output mode, that outputs RelEx relations in a format similar to that of the Stanford parser.   More commonly used, however, is the output format that produces OpenCog "Atomese" that can be loaded into the OpenCog system for further processing, including transformation into logical relationships via the RelEx2Logic system.

\subsubsection{RelEx2Logic}

The core idea of RelEx2Logic, operationally, is to map RelEx relationships into logical semantic interpretations (i.e. PLN logical expressions) via applying a core set of simple rewrite rules, along with more complex post-processing rules (which deal with various aspects of complex sentences, and also with semantic issues such as the concept-instance distinction). 

The representation of RelEx2Logic rules utilizes OpenCog's Atomspace representation \cite{Goertzel2014b}, in which various formal systems are given common expression as weighted, labeled hypergraphs.  In the Atomspace representation, both link parses, PLN predicate-logic expressions and type expressions are represented as hypergraphs with particular node and link types, which are reviewed e.g. in \cite{Goertzel2014a,  Goertzel2014b} \footnote{ see also \url{http://wiki.opencog.org/w/AtomSpace}}.   PLN's truth value objects are then represented as weights on certain nodes and links; and may alternately be represented as hypergraphs themselves (since e.g. there are NumberNodes representing numbers, and ListLinks representing links, so that a number-list like $(.9,5)$ can itself be represented as a node-and-link structure).

Each of the core RelEx2Logic rewrite rules takes as input a subgraph of a syntactic parse graph (representing RelEx output) satisfying certain constraints, and outputs an Atom hypergraph.  In practice the rules required generally take the form of pairs $(G,A)$, where 

\begin{itemize}
\item $G$ is a graph whose nodes are either words or variables, and whose links are RelEx relationship types
\item $A$ is a hypergraph whose nodes are either words, variables or special linguistic nodes (drawn from a  small vocabulary of such), and whose hyper-edges are OpenCog Atom types (e.g. InheritanceLink, EvaluationLink).  
\item The lists of variables in $G$ and $A$ must be the same
\end{itemize}

\noindent In a RelEx2Logic context, rules matching this description are called "simple mapping rules."   

For the simple mapping rules actually needed for handling human language, the constraint that {\bf each edge in $G$ maps into a single hyper-edge in $A$} appears to hold true.   Mathematically, this latter constraint implies that each of the rewrite rules is individually a {\bf graph homomorphism} \cite{Voloshin2009}, which then implies that a collection of rewrite rules applied together is also a graph homomorphism.  Categorically, this observation is important because it makes it simple to consider the mapping from link parses to PLN expressions via RelEx and RelEx2Logic as a morphism between the category corresponding to link parses and the category corresponding to PLN expressions.

A simple example of such a rule is $(G,A)$ where

$$
G = \{ \_subj(v_1,v_2), \_obj(v_1, v_3) \}
$$

$$
A = (EvaluationLink \ v_1 \ v_2 \ v_3)
$$

\noindent This maps a verb $v_1$ with subject $v_2$ and object $v_3$ into an OpenCog EvaluationLink with $v_1$ as the predicate and $(v_2, v_3)$ as the argument list.  Of course, most rules are more complex than this; see \url{http://wiki.opencog.org/w/RelEx2Logic_rules} for the rule-based in use in early 2016.

The application of this simple example mapping rule to  "The cat chased a snake" yields the primary output \footnote{For a formal explication of the OpenCog Atomese notation used in this and further examples, see \cite{Goertzel2014a}}

{\tt\begin{small}\begin{lstlisting}
EvaluationLink 
   PredicateNode chase@3453432
   ListLink
      ConceptNode cat@1243546464
      ConceptNode snake@564636322
\end{lstlisting}\end{small}}

\noindent Here e.g. "chase@3453432" refers to a specific instance of the general predicate "chase".  

This "primary output"  is only a small fraction of the total set of Atoms created in the OpenCog Atomspace upon interpretation of this sentence; but it communicates the crux of what's going on.   In more standard notation, what this means is just

{\tt\begin{small}\begin{lstlisting}
chase@3453432(cat@1243546464, snake@564636322)
\end{lstlisting}\end{small}}

\noindent -- but the OpenCog Atomspace hypergraph notation is more explicit, expanding the evaluation relationship into its own hypergraph link, and explicitly identifying the type of each entity.

\subsubsection{A Type-Theoretic View of Syntax to Semantic Mapping}

One may also view this sort of mapping in terms of type theory.   For instance, if we map "dogs bark" into

{\tt\begin{small}\begin{lstlisting}
EvaluationLink
   PredicateNode "bark"
   ConceptNode "dogs"
\end{lstlisting}\end{small}}

\noindent then we are in effect assigning "dogs" to the type $T$ corresponding to

{\tt\begin{small}\begin{lstlisting}
EvaluationLink
   PredicateNode "bark"
   ConceptNode $X
\end{lstlisting}\end{small}}

\noindent with free variable $\$X$ .

Categorically, the arrow

{\tt\begin{small}\begin{lstlisting}
(ConceptNode $X) 
----> 
(EvaluationLink (PredicateNode "bark") (ConceptNode $X))
\end{lstlisting}\end{small}}

\noindent corresponds to this type expression.

The symmetry of the algebra of logic here consists in the fact that it is not "dog as it
connects to the left" or "dog as it connects to the right" that
belongs to this type $T$, it is just plain old "dog".

\paragraph{Dependent Types and Anaphora}.  Dependent types come in here when one has semantic mappings that are
left unresolved at the time of mapping.   Anaphora are a standard
example, for instance sentences like "Every man who owns a donkey,
beats it" \cite{Bekki2014} 

The point here is that we want to give "beat" a type of the form 

{\tt\begin{small}\begin{lstlisting}
EvaluationLink
   PredicateNode "beat"
   ListLink
      SomeType $y
      SomeType $z
\end{lstlisting}\end{small}}

\noindent where then the type of $\$y$ is the type of men who own donkeys, and the
type of $\$z$ is the type of donkeys.  But the rule for parsing "beats
it" should be generic and not depend on the specific types of the
subject and object of "beats," which will be different in different
cases.

In accordance with the above ideas, 
the output of this semantic mapping framework, given a sentence, can
be viewed as a set of type judgments, i.e. a set of assignations of
terms to types.    \footnote{Recall that categorically, assigning term $t$ to type $T$
corresponds to an arrow $t \circ ! : \Gamma \rightarrow T$ where $!$ is an arrow
pointing to the unit of the category and $\Gamma$ is the set of type
definitions of the typed lambda calculus in question, and $\circ$ is
function composition.}

The case of "Every man who has a donkey, beats it" illustrates that in
order to get compositionality for formally odd but commonsensically natural sentences
like this, you want to have dependent types in your lambda calculus at the
logic end of your mapping.

\subsubsection{Reconceptualizing Linguistic Compositionality}

One philosophically nice observation here is that: In this framework,
Frege's "principle of
compositionality" corresponds to the observation that there is a
morphism from the asymmetric monoidal category corresponding to link
grammar, into the symmetric locally cartesian closed category
corresponding to lambda calculus with dependent types.

This principle means that, in a certain sense, you can get the meaning of a linguistic
whole (e.g. a sentence) by combining the meanings of the linguistic parts (e.g. the
words or phrases).  However, this must be interpreted carefully.   Of course, the
set of mental associations of a combination of several linguistic parts, may very prominently
feature some entity, even if that entity exists only dimly as a mental association of any
of the individual parts.   However, nevertheless, for something to exist as an association
of a linguistic combination, this thing must exist {\it at least dimly} as an association of
each of the parts.   

The key subtlety here has to do with the propagation of probability values (e.g. probabilities of association between linguistic and nonlinguistic structures) through the algebra of logic, as reviewed extensively in \cite{Goertzel2008}.  Linguistic structure
maps into crisp logical structure, and sometimes provides guidance as to probability values to be attached to logical terms and relationships; but cognitive associations of logical terms and relationships often have much finer grained probability values, which then propagate through linguistic parts and wholes, sometimes giving the logical cognate of a a linguistic whole a probability value that cannot be calculated from the probability values of the logical cognate of the linguistic parts, without consideration of a great deal of additional cognitive content.   Furthermore, In many cases,
two statements that are equivalent in crisp logic, may end up with different probabilistic truth values if one considers probability values attached to the  atomic entities in the statement, and then uses heuristics to estimate the probability value of the overall statement -- because of imprecision in the heuristics involved.

\subsection{Generation as Reverse Comprehension}

One elegant effect of formalizing language comprehension as we have done, is that it becomes
straightforward to model language generation as the reverse of language comprehension.

At the highest level, language generation  may be broken down into three stages: macroplanning, microplanning and surface realization \cite{Goertzel2014a, Goertzel2014b, Jurafsky2009, Reiter2000}.   Macroplanning is overall discourse management; microplanning deals with breaking down fairly small-sized chunks of semantic content into sentence, inserting anaphora, word choice, and so forth.   Surface realization deals with mapping semantic structures into syntactic structures, where in the overall generation pipeline it is fed semantic structures chosen and formed by macroplanning and microplanning.   It is surface realization that we will model here as the reverse of the syntax-to-semantics mapping operation that occurs in language comprehensions.

In the OpenCog system, the SuReal surface-realizer component takes small collections of Atoms, selected by the OpenCog microplanner, and transforms them into Atoms corresponding to English sentences to be uttered to a human conversation partner or application user.

Very broadly speaking, what the SuReal algorithm (described in \cite{Lian2010}, with its previous name SegSim, and with attention to its cognitive neuroscience motivations; and described in a more current way in \footnote {\url{https://github.com/opencog/opencog/tree/master//opencog/nlp/sureal}} ) is doing is simply to reverse the various graph and hypergraph rewrite rules described in the previous sections on RelEx and RelEx2Logic.   However, this is not entirely simple, because the rules create homomorphisms rather than isomorphisms.  Any one Atom structure may be produced by many different link-grammar structures, because there are many grammatical ways to produce any given idea.   But not all the grammatical structures corresponding to different subsets of a given Atom set needing articulation, will necessarily be grammatically compatible with each other.  So one has a constraint satisfaction problem, which in general will have multiple solutions, with varying levels of syntactic ambiguity and subjective human naturalness.   The SuReal algorithm represents a heuristic approach to this problem.

More precisely, suppose we have an Atom set $A = \{A_i\}$; and let $R = \{ R_{j} \}$ denote the set of all graph rewrite rules $R_j$ with the property that $R_j$ maps at least one link parse subtree into some nonempty subset of $\{A_i\}$.  Let $R^i \in R$ denote the set of rewrite rules that produce an Atom set including the particular Atom $A_i$; we may write $R^i = \{R^i_k\}$, with $R = \bigcup_i R^i$.   Let $m_g(r )$ denote the proposition that the rewrite rule $r$ matches some subgraph of the graph $g$.

Given this set-up, the problem of generating a sentence expressing the Atom set $A$ boils down to finding some link parse $g$ that 

\begin{itemize}
\item parses correctly according to link grammar
\item satisfies the expressiveness condition

$$
\bigwedge_i \bigvee_k m_g(R^i_k),
$$

\item satisfies an assumed "aesthetic condition", initially: that it would either not parse or not satisfy the expressiveness condition if any of its words were removed
\end{itemize}

\noindent Given a link parse $g$, producing the relevant sentence is trivial.  The task of generating a sentence-set expressing $A$ reduces to choosing a way to partition $A$ into subsets, so that each can be acceptably expressed via a single sentence.  

\paragraph{From Language to Logic and Back Again}   In sum, we have seen that significant aspects of natural language comprehension can be modeled in terms of mappings from syntax into logical semantics, which can be viewed in terms of morphisms from the asymmetric monoidal category of grammar (as modeled in pregroup grammar, which is equivalent to link grammar), into the symmetric locally closed Cartesian category of typed lambda calculus,  which is the core of logical inference (and if one adds appropriately quantified uncertain truth values, provides an extension of intuitionistic logic).  Significant aspects of natural language generation can be viewed in terms of tracing these morphisms in the opposite direction; though other aspects, such as pragmatics oriented micro and macro planning, need to be dealt with using broader sorts of inference.

\section{A Categorial View of Spatiotemporal Perception}

The cognitive value of language lies largely in its ability to reflect the structure of non-linguistic domains of experience, in ways that are amenable to flexible manipulation and communication.  This ability can be modeled via constructing mappings between these non-linguistic domains and logic, which then via composition automatically results in mappings between these nonlinguistic domains and language.   Here we illustrate this point via the example of spatial structure, a relatively simple but highly important nonlinguistic domain.

The mathematical structures used to represent syntactic structure in pregroup and link grammar, can with slight extension be used to represent spatiotemporal structure in sensory data as well.   The key concept here is that of an "image grammar" \cite{Noden2011} -- a system of habitual arrangements of visual forms.  For instance, an image grammar regarding human faces would contain probabilistic rules to the effect that "an eye usually occurs next to another eye", "eyes normally occur above a nose", etc.  The notion of an image grammar can be extended to that of a "3D movie grammar", i.e. a system of habitual arrangements of 3D patterns that may change over time in specific ways.

One can then look at morphisms from the product of these inferred image grammars, into the typed lambda calculus.   These morphisms map the grammar of spatiotemporal forms, into the algebra of logic.  Depending on the spatial grammar relationships involved, this may provide a significant degree of simplification and normalization.  For instance the logical relations $near(x,y)$ or $next\_to(x,y)$ each summarize a variety of different particular spatial relationships.

In the following we explore some restricted cases in which the formalization of these ideas is especially concise.

\subsection{Region Connection Calculus}

To make our discussion of the mapping between linguistic and spatial structure maximally concrete, we will initially focus attention on a small set of spatial relationships that are particularly straightforward to formalize.   After treating the issues in this context, we will then generalize.

We will first consider general-purpose relationships between spatial regions.   One can straightforwardly model
three types of relationships between spatial regions: topological,
directional and metric.   Here we will focus on the topological aspect.

\begin{figure}[H]
\begin{center}
\ifpdf
\includegraphics[scale=0.65]{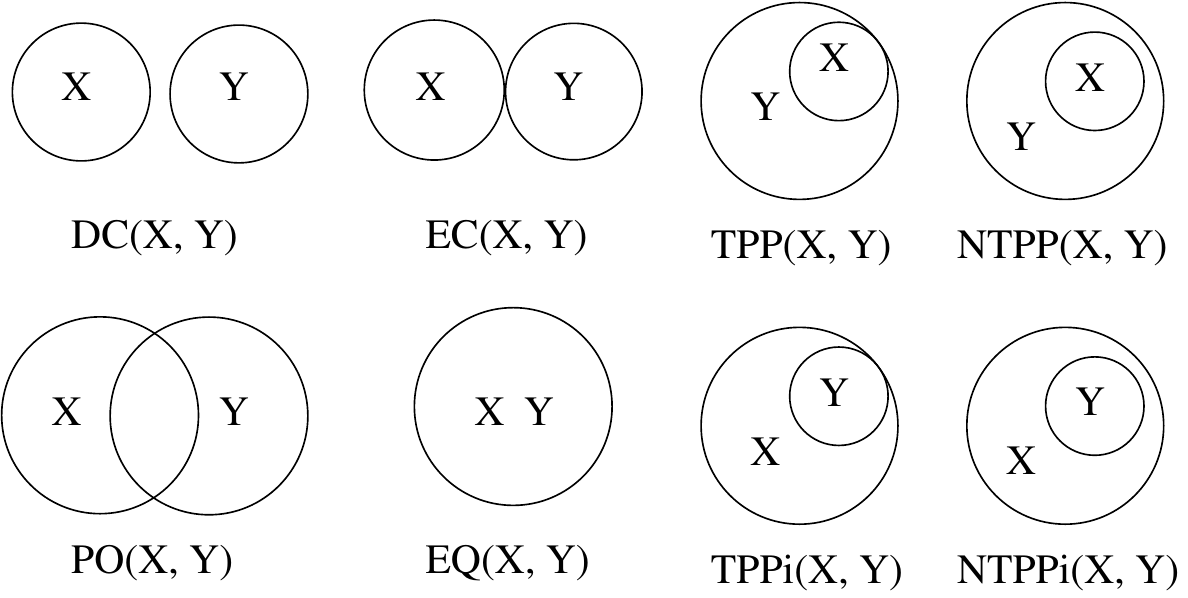}
\else
\includegraphics[scale=0.65]{RCC8.eps}
\fi
\caption{\label{fig:RCC8}The eight base relationships of RCC-8}
\end{center}
\end{figure}

The most popular calculus dealing with topology is
the Region Connection Calculus (RCC) \cite{RanCuiCoh93},
relying on a base relationship $\CR$ (for $\CRl$)
and building up other relationships from it, like $\PR$ (for
$\PRl$), or $\OR$ (for $\ORl$).
For instance $\PR(X,Y)$, meaning $X$ is a part of $Y$,
can be defined using $\CR$ as follows

$$
\PR(X,Y)  \textrm{ iff }  \forall Z \in \Un, \CR(Z,X) \implies \CR(Z,Y)
$$

\noindent where $\Un$ is the universe of regions. RCC-8 models eight base relationships,
see Figure \ref{fig:RCC8}.

It is also possible, using the notion of convexity, to model more
relationships such as inside, partially-inside
and outside; this is done in RCC-23 \cite{Bennett94}.  
RCC-3D extends the RCC framework to relationships between 3D regions \cite{Albath2010}.
An extension to simple closed regions of 4D spacetime can be constructed straightforwardly
as well; the mathematical relationships involved are essentially independent of dimension.

\subsection{Allen Interval Algebra}

\begin{figure}[H]
\begin{center}
\includegraphics[width=10cm]{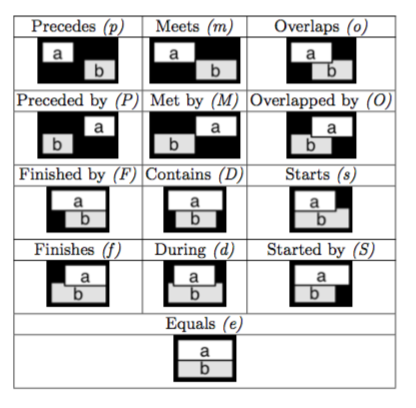}
\caption{Temporal relations in Interval Algebra}
\label{fig:allen}
\end{center}
\end{figure}

The RCC can be viewed as an extension to higher dimensions of the Allen interval algebra, which describes the simple spatial relationships between 1D closed intervals.  Figure \ref{fig:allen} illustrates the relationships characterizing Allen interval algebra; these are crisp in nature, but uncertain variations have also been articulated \cite{GerBev02}.

Conceptually, the Allen interval algebra relations can be viewed as analogous to RelEx relationships.   They are symmetric, even though the direct relations between specific intervals in the 1D line have an obvious left-right asymmetry (just as the symmetry of RelEx relations abstracts away the asymmetry of word positions in natural language productions).  

The relationship between linguistic grammar and spatiotemporal grammar may be particularly easy to see in the context of Allen interval algebra.   Once can view the Allen interval algebra relationships as abstractions of an underlying link grammar type layer, whose link types correspond to pairs of the form (object category, basic interval relationship).   The basic interval relationships needed are three in number: "overlaps", "abuts", "gap between" (O, A and G if we wish).   

\subsection{Basing Image Grammars on RCC and Interval Algebra}

We can use RCC and Allen interval algebra to formulate a type of image grammar, patterned on the link grammar framework, where each type of visual entity in the image grammar gets a dictionary entry composed of connectors of various types.  Each connector type is constructed based on joining an RCC/Allen relationship type with a unique identifier associated with the visual entity type, and optionally with a coordinate-axis type (e.g. horizontal or vertical or front/back).  To have a concrete syntax, we can define a connector type label as having one of the forms
$$
\textrm{entity-type-identifier\_axis-identifier\_topological-relation-identifier\_signum} 
$$ 
\noindent (where {\it signum} is either + or -).

So for instance, if we want to say that an eye should have another eye to either the left or to the right of it, we would say that the image grammar dictionary entry for the visual entity type Eye has a disjunct of the form 
$$
\textrm{eye\_h\_G+} \vee \textrm{eye\_h\_G-}
$$ 
\noindent  Here the entity type is "eye", the coordinate axis identifier is $h$ (horizontal), the topological relationship identifiers is $G$ ("gap between" from Allen interval algebra), and the directionality of the relationship may be either right (+) or left (-).   This means that each instance of Eye must have another instance of Eye involved in a "gap between" ($G$) relationship with it, either to the right or to the left, along the horizontal ($h$) axis.   

Supposing there must be a nose below an eye, we could then also say that in the dictionary entry corresponding to Eye, there would be a disjunct of the form 
$$
\textrm{eye-nose\_v\_G-}
$$
\noindent (a link of type $\textrm{eye-nose\_v\_G}$, meaning "from eye to nose", extending to the right along the vertical axis, meaning down), whereas in the dictionary entry corresponding to Nose, there would be a disjunct of the form 
$$
\textrm{eye-nose\_v\_G+}
$$
\noindent (a link of type $\textrm{eye-nose\_v\_G}$ extending to the left along the vertical axis, meaning up).  A more complex dictionary entry for Eye could specify 
$$
(\textrm{eye\_h\_G+} \wedge \textrm{eye-nose\_v\_G-} \wedge \textrm{eye-nose\_h\_G+}) \vee (\textrm{eye\_h\_G-} \wedge \textrm{eye-nose\_v\_G-} \wedge \textrm{eye-nose\_h\_G-})
$$
\noindent, which specifies that if the Eye is a left eye, then the nose occurs to its right; and if the Eye is a right eye, then the nose occurs to its left; but in either case, the nose occurs below the eye.   De Morgan's identities let us expand this into conjunctive normal form so it looks like an ordinary link grammar entry.

\subsection{Inferring Image Grammars from Deep Learning Networks}

In a prior paper \cite{Goertzel2013}, one of the authors and his colleagues have shown it is possible to infer image grammars from images by way of deep learning computer vision systems.   Specifically, in this work, a frequent subtree miner is used to recognize probabilistically significant patterns in the states of the DeSTIN deep neural net algorithm applied to recognize patterns in 2D images.   \footnote{While the formalism of image grammar was not explicitly used in this prior paper, the geometric relations inferred among these frequent patterns were clearly equivalent to an image grammar.}

In the DeSTIN system used there, the 2D input image is partitioned into a quadtree, with each quadtree region corresponding to a "DeSTIN node."  Each of these nodes contains K clusters, where the clusters comprise elements of a local pattern dictionary, where the dictionary is learned dynamically as part of the DeSTIN algorithm.   The state of the DeSTIN network at each point in time then consists of a quadtree with each node labeled with an integer in the range $1 \ldots K$ (the most prominent pattern from the dictionary in that node at that point in time).   

Frequent patterns among these labeled quadtrees are then conjunctions such as "label 4 at node (1,1) on layer 1 AND label 5 at node (1,2) on layer 2".   For instance, if DeSTIN is fed characters, these frequent patterns will correspond to common geometric components of characters, such as ascenders or curves.  The common relations among these geometric components form an image grammar defining the structure of characters, including grammatical rules corresponding to observations like "a vertical ascender sometimes occurs (horizontally) adjacent to a circle"  In standard OpenCog Atomese notation, this would be written as a relation like:

{\tt\begin{small}\begin{lstlisting}
EvaluationLink <.5,.9>
   PredicateNode "adjacent"
   ListLink
      ConceptNode "pattern_2
      ConceptNode "pattern_7"
\end{lstlisting}\end{small}}

\noindent where $\textrm{pattern\_2}$ refers to "vertical ascender" and $\textrm{pattern\_7}$ refers to semicircle.  Each of these ConceptNodes would then be linked to an AndLink joining PredicateNodes describing the labels at various nodes of the DeSTIN hierarchy.   In the link grammar notation used here, this would instead be notated by putting
$$
\textrm{pattern-7\_h\_EC+} \vee \textrm{pattern-7\_h\_EC-}
$$ 
\noindent in the dictionary entry of $\textrm{pattern-2}$ (using the EC label for "adjacency", from the RCC); and putting
$$
\textrm{pattern-2\_h\_EC+} \vee \textrm{pattern-2\_h\_EC-}
$$ 
\noindent in the dictionary entry of $\textrm{pattern-7}$.

The same basic approach can be taken if one has a DeSTIN-like deep neural network corresponding to 3D space, or 3D space + 1D time.   For instance, in a 3D implementation of the DeSTIN algorithm \cite{Arel2009}, there is a "node" corresponding to each node in an octree tiling a certain region of 3D spacetime.    If one replaces the clustering in the standard DeSTIN with other algorithms such as denoising autoencoders or other neural networks (a strategy we are taking in our current research work), the logic regarding pattern mining and image grammar remains the same.  

\subsection{Connection with Symbolic Dynamics}

Extending the ideas from \cite{Goertzel2013} a little further, patterns in the states of deep neural networks may be viewed as a special case of something more general and subtle.  Recall "symbolic dynamics" \cite{Morse1938}  -- wherein one can discretize the state space of any dynamical system, and then represent each of its trajectories as a word in the language for which the discretization-cells of the state space are the elements of the alphabet.   Using this tool one can, for example, map a recurrent neural network into a symbolic system via expressing the trajectories of the network (during fixed, e.g. equally-spaced, time-intervals) as expressions in a formal grammar.

In general, however, figuring out the right way to discretize the states of a general recurrent neural network, so as to get the simplest and most inference-amenable grammar from the symbolic dynamics, is a tough problem.   Having a specific structure to one's neural network, as is the case with deep neural nets whose structure reflects that of regions of spacetime, makes things simpler because it tells one how to discretize the state space

In a recurrent deep neural network with spatiotemporal structure and significant top-down as well as bottom-up information flow (such as a DeSTIN network recognizing patterns in videos and incorporating feedback connections from parent nodes to child nodes), a recurring pattern of activity across multiple regions in the network will often be be a {\it persistent transient} in the system's dynamics (in some cases it could be a strange attractor or, say, a terminal attractor \cite{Zak1989}).   So identifying the discretization-cells of the state-space of the network, with combinations of {\it discretized-states of regions of the network corresponding to localized regions of spacetime}, will often work and capture most of the cognitively important structure.  The partitioning problem is solved by the network's design.

\section{Symbol Grounding as Chaining of Morphisms}

Given the mappings from natural language to logic, and spatiotemporal structure to logic, that we have outlined above, it is now straightforward to articulate a mapping from linguistic to spatiotemporal structure (and back) via logic.   We will then see afterwards how a similar approach can be taken with actions, parallel to our treatment of visual perception.

A key feature of this mapping is the more symmetric nature of the structure of logic, as compared to the more fundamentally asymmetric structures of linguistic and spatiotemporal relationships.   The mappings from language to logic, and spacetime structure to logic, involve reducing concision while increasing symmetry.   In the more symmetric realm of logic, parallels between linguistic and spatiotemporal structure are easier to represent.  

\subsection{A Specific Example: Partial Overlap}

To understand the relationships involved here fully, let's take a look at the example of the RCC
"partially overlaps" relationship, $PO(x,y)$.   In English the straightforward way to say this is "x partially overlaps y".  
Of course there are more idiomatic ways of saying this, and there are also complexities related to ambiguity --
e.g. "x overlaps y" would normally be understood to imply "x partially overlaps y" to a certain probabilistic degree.
But for illustrative purposes we can stick with the simplest case.

The link parse for "x partially overlaps y" looks like (inserting the concrete words "Jack" and "Jill" for x and y, for simplicity).

{\tt\begin{small}\begin{verbatim}
    +-------------------Xp-------------------+
    +------------>WV----------->+            |
    |        +--------Ss--------+            |
    +---Wd---+       +-----E----+---Os---+   |
    |        |       |          |        |   |
LEFT-WALL Jack.b partially overlaps.v Jill.f . 
\end{verbatim}\end{small}}

\noindent and the link parse for "y is partially overlapped by x" looks like

{\tt\begin{small}\begin{verbatim}
    +-------------------------Xp-------------------------+
    +--------------->WV--------------->+                 |
    |              +---------Pv--------+                 |
    +---Wd---+--Ss-+      +------E-----+---MVp--+-Js-+   |
    |        |     |      |            |        |    |   |
LEFT-WALL Jill.f is.v partially overlapped.v-d by Jack.b . 
\end{verbatim}\end{small}}

\noindent The RelEx output for these two cases is identical, 

{\tt\begin{small}\begin{verbatim}
Dependency relations:

    _obj(overlap, Jill)
    _advmod(overlap, partially)
    _subj(overlap, Jack)

Attributes:

    tense(overlap, present)
    penn-POS(overlap, VBZ)
    pos(overlap, verb)
    pos(., punctuation)
    gender(Jill, feminine)
    definite-FLAG(Jill, T)
    penn-POS(Jill, NN)
    person-FLAG(Jill, T)
    pos(Jill, noun)
    noun_number(Jill, singular)
    penn-POS(partially, RB)
    pos(partially, adv)
    gender(Jack, person)
    definite-FLAG(Jack, T)
    penn-POS(Jack, NN)
    person-FLAG(Jack, T)
    pos(Jack, noun)
    noun_number(Jack, singular)
\end{verbatim}\end{small}}

\noindent as is the RelEx2Logic output, 

{\tt\begin{small}\begin{lstlisting}
InheritanceLink
   ConceptNode "partially@196c"
   ConceptNode "partially"
ImplicationLink
   PredicateNode "overlaps@5d9c"
   PredicateNode "overlap"
InheritanceLink
   SatisfyingSetLink
      PredicateNode "overlaps@5d9c"
   ConceptNode "partially@196c"
InheritanceLink
   ConceptNode "Jill@00cd"
   ConceptNode "Jill"
EvaluationLink
   DefinedLinguisticPredicateNode "definite"
   ListLink
      ConceptNode "Jill@00cd"
InheritanceLink
   ConceptNode "Jack@6725"
   ConceptNode "Jack"
EvaluationLink
   DefinedLinguisticPredicateNode "definite"
   ListLink
      ConceptNode "Jack@6725"
EvaluationLink
   PredicateNode "overlaps@5d9c"
   ListLink
      ConceptNode "Jack@6725"
      ConceptNode "Jill@00cd"
EvaluationLink
   PredicateNode "overlaps@5d9c"
   ListLink
      ConceptNode "Jack@6725"
InheritanceLink
   InterpretationNode "sentence@60e5_parse_0_interpretation_$X"
   DefinedLinguisticConceptNode "DeclarativeSpeechAct"
InheritanceLink
   SpecificEntityNode "Jill@00cd"
   DefinedLinguisticConceptNode "female"
InheritanceLink
   SpecificEntityNode "Jill@00cd"
   ConceptNode "Jill"
InheritanceLink
   PredicateNode "overlaps@5d9c"
   DefinedLinguisticConceptNode "present"
\end{lstlisting}\end{small}}

On the other hand, the direct rendering of the RCC relationships $PO(x,y)$ into Atomese looks like

{\tt\begin{small}\begin{lstlisting}
EvaluationLink
   PredicateNode "RCC_Partial_Overlap"
   ListLink
      ConceptNode $x
      ConceptNode $y
\end{lstlisting}\end{small}}

\noindent and it is easy to see that the equivalence

{\tt\begin{small}\begin{lstlisting}
LambdaLink
   VariableNode $x, $y
   EquivalenceLink <s,c>
	  ANDLink
	     EvaluationLink
			PredicateNode "overlaps"
			   ListLink
			      $x
			      $y
		 InheritanceLink
			SatisfyingSetLink
			   PredicateNode "overlaps"
			   ConceptNode "partially"
   EquivalenceLink
 	  EvaluationLink
 	     PredicateNode "RCC_Partial_Overlap"
 	        ListLink
 	           $x
 	           $y
 \end{lstlisting}\end{small}}

\noindent isomorphically maps the $\textrm{RCC\_partial\_overlap}$ relation into the RelEx2Logic output from the
linguistic form "partially overlaps."  

Note that this EquivalenceLink uses e.g.

{\tt\begin{small}\begin{lstlisting}
InheritanceLink
   SatisfyingSetLink
      PredicateNode "overlaps"
   ConceptNode "partially"
 \end{lstlisting}\end{small}}
 				
\noindent whereas the RelEx2Logic output cited uses

{\tt\begin{small}\begin{lstlisting}
InheritanceLink
   ConceptNode "partially@196c"
   ConceptNode "partially"
ImplicationLink
   PredicateNode "overlaps@5d9c"
   PredicateNode "overlap"
InheritanceLink
   SatisfyingSetLink
   PredicateNode "overlaps@5d9c"
   ConceptNode "partially@196c"
 \end{lstlisting}\end{small}}		
 
 \noindent However, the former follows from the latter via a few simple and obvious steps of inference, which the PLN logic engine
 can carry out unproblematically; so it is fair to consider the former as directly implied by the RelEx2Logic output.   If one does a minimal
 amount of forward-chaining PLN inference on the RelEx2Logic output, when running OpenCog, then one will obtain the former as part
 of the resulting Atom-set.	

If the truth value on the EquivalenceLink is $<s,c> = <1,1>$, then we have a pure mathematical tautology with a truth value of unity.   On the other hand, if there is some ambiguity on the natural language side of the equivalence, then the truth value may be less than unity in strength $s$ or confidence $c$.   This would be the case if there was a possibility of the construct headed by the ANDLink actually having some other interpretation.  In that case, the mapping would be correct, but only under one of the possible interpretations of the ambiguous natural language predicates "overlaps" and "partially."   

In general, the RelEx2Logic subsystem, in OpenCog, attempts to remove all ambiguity regarding the {\it logical structure} of a natural language utterance, i.e. meaning the shapes of the Atomspace hypergraphs representing its interpretation, and the labels of the nodes in these hypergraphs.   However, it does not seek to remove all ambiguity regarding the meanings of the labels of the nodes in the Atomspace hypergraphs it outputs.   Reducing and otherwise managing this kind of ambiguity is also important for linguistic cognition, but in the OpenCog framework is intended to be handled via other processes such as PLN inference based on data induced from language and experience.

\subsection{RCC Maps Morphically onto a Subset of English}

One clear and simple way to look at the morphism between RCC and natural language is to consider a subset of English consisting of the words in the RCC descriptors

{\tt\begin{small}\begin{verbatim}
disconnected 
externally connected 
equal 
partially overlapping 
tangential proper part 
non-tangential proper part 
\end{verbatim}\end{small}}

\noindent , plus the word "region," the positive integers, and the connecting words

{\tt\begin{small}\begin{verbatim}
is
with
from
to
a
not
and
or
\end{verbatim}\end{small}}

Suppose we call this subset of English "Minimal RCC English."   In Minimal RCC English, one can say things like

{\tt\begin{small}\begin{verbatim}
Region 5 is a non-tangential proper part of Region 7
Region 7 is not equal to Region 6
Region 6 is partially overlapping with Region 55
Region 3 is equal to Region 7, 
              and Region 7 is externally connected with Region 9
Region 4 is equal to Region 5, and Region 5 is equal to Region 6
etc.
\end{verbatim}\end{small}}

The link grammar dictionary entries corresponding to the words in Minimal RCC English, and required to parse sentences in Minimal RCC English, may collectively be called "Minimal RCC English Link Grammar."   This does not include, for instance, entries in the dictionary entry for "with" that are used only to connect "with" with words {\it not} in Minimal RCC English.

It should be clear from the foregoing discussion that

\begin{myproposition}
There is a morphism between the asymmetric monoidal category corresponding to the Minimal RCC English Link Grammar, and the closed cartesian category corresponding to the set of propositional relationships in the RCC itself.
\end{myproposition} 

\noindent Obviously, the English articulation of a spatial relationship between two entities has more communicative value than the corresponding propositional RCC formulation, in many contexts.   However, the formal RCC version has the advantage of transparent connection to various inference rules, such as the RCC multiplication table, as well as basic logical deduction which lets one conclude e.g. that Region 4 is equal to Region 5, and Region 5 is equal to Region 6" implies "Region 5 is equal to Region 6."

The Minimal RCC English Link Grammar is a subset of overall English link grammar, and the RCC's propositional relationships can straightforwardly be embedded in PLN (or any other moderately powerful logic), and hence constitute a subset of the logic used by OpenCog or other AI logic formulations such as FCG or Cyc.   This proposition thus can be seen as formalizing a small subset of the broader morphism between linguistic and logical structure.   

To see how one might extend this basic proposition bit by bit, suppose we added to our Minimal RCC English Link grammar: 

\begin{itemize}
\item a set of nouns and adjectives for identifying visual entity types
\item the words "on" , "the", "vertical", "horizontal", "axis" , "vertically", "horizontally", "related"
\item terms for the Allen Interval Algebra relationships
\item "or" and "and" (to allow specification of multiple relationships associated with the same entity type)
\item a few uncertain qualifiers: "often", "usually", "occasionally"
\end{itemize}

We may call this the Extended RCC/Allen English Link Grammar.  Using this slightly extended grammar, we can articulate sentences such as

{\tt\begin{small}\begin{verbatim}
A semicircle is often externally connected with  
a vertical ascender, on the horizontal axis.

An eye is usually vertically related to a nose 
with a gap between, or horizontally related to 
a nose with a gap between
\end{verbatim}\end{small}}

In this extended grammar, we can formulate English statements corresponding to the concrete image grammar examples given above; and one has the relationship

\begin{myproposition}
There is a morphism between the asymmetric monoidal category corresponding to the Extended RCC/Allen English Link Grammar, and the closed cartesian category corresponding to: The set of propositional relationships between atomic terms, where each relationship has the form $(\textrm{vertical} \vee \textrm{horizontal}, \textrm{right} \vee \textrm{left}, R)$, where $R$ is a label for either an RCC or Allen interval algebra relationship.    There is also a morphism between this closed cartesian category, and the asymmetric monoidal category corresponding to the image grammar outlined above, comprising a link grammar with link types of the form 
$$
\textrm{entity-type-identifier\_axis-identifier\_topological-relation-identifier\_signum}
$$
\end{myproposition} 

The limited subsets of English grammar we have identified above are both highly constrained semantically, and highly awkward syntactically.   It is clearly straightforward to extend them to include more colloquial English, as well as more types of spatial relationships, including geometric as well as topological relationships, for example.   In general, the broader morphism between linguistic grammar and image grammar can be approximated via formulating additional propositions similar to the above encompassing larger and larger subsets of English.   It is not clear how useful the explicit articulation of such propositions is; however, we present the example proposition above in order to concretely exemplify the type of morphism between language and perception we are referring to.

%%%%%%%%%%%%%%%%%%%%%%%%%%%%%%%%%%%%%%%%%%%%%%%%%%%%%%%%%%%%%%%%%%%%%%%%%%%%

\section{Modeling Actions}

Language may be grounded in actions as well as perceptions.   To formalize this one may introduce "action grammars", similar in concept to image grammars.

In the action context, one does not have a 4D spatiotemporal structure, but rather has a higher-dimensional structure referred to as a configuration space or C-space \cite{Morales2006}.  For instance in a robot with 50 motors, one might have a 150 dimensional space corresponding to the 3 degrees of freedom of each motor.  A specific movement corresponds to a trajectory over time in configuration space.   

There are many different ways to model complex movements; here we will briefly describe a hierarchical and compositional approach, which leads naturally to a grammatical model of movement.  In this approach, a complex movement is represented as a combination of simpler movements, where each movement corresponds to a trajectory over time in some subspace of C-space.   

A movement-type or "animation" corresponds to a set of movements, parametrized by a certain set of $k$ (normally real number) parameters.   In the simplest case these parameters are each restricted to some interval.   In a more complex case, there is some recognizer function, mapping $R^k$ into ${0,1}$, indicating when a $k-tuple$ is a legal parameter vector for that movement type.  Example animations would be a walking gait, a running gait, a "reaching" arm movement, etc.

Many animations are composed via combining simpler animations corresponding to subsets of actuators.   So e.g. a reaching arm animation may be decomposed into a "shoulder + elbow reaching" animation and a "wrist + hand reaching" animation.   A "hand reaching" animation may be decomposed into a number of "finger reaching" animations.   Much as an object-type or object-part in vision corresponds to a pattern of organization of percepts that is observed repeatedly; similarly, an animation corresponds to a pattern of organization of actuator movements that is enacted repeatedly.

Furthermore, the set of parameter-vectors allowed for a particular animation, may often be partitioned naturally into a collection of discrete cells.   For instance, one can reach forward, backward, up or to the side.   Martial arts disciplines categorize "kicks" into a refined collection of discrete categories.   These cells may be considered to constitute animations in themselves (in this case one has multiple animations regarding the same set of actuators).

In the hierarchical model we are pursuing here, a specific movement is identified with a set of animations, each beginning at a particular time-offset from the initiation of the movement.  At any given moment in time, a number of animations may be current.  The grammar of these animations, then, resembles a natural language in having a natural sequential order, but differs in terms of the presence of concurrency.   Movement grammar is much like the grammar one would expect  to find in the language of a creature with a large number of mouths, which could easily utter multiple words at the same time.  

To model movement grammar using link grammar formalism, one can introduce link types of the form $A\_r{\it signum}$ where $A$ is an animation label, and $r$ is an Allen interval algebra relationship.   So for instance, to express the fact that moving one's lower leg back is often the first part of a "kick forward" action, one would add the entry 
$$
\textrm{kick-lower-leg-forward}\_a+
$$
\noindent to the dictionary entry for$\textrm{move-lower-leg-back}$.   This indicates that a 
$$
\textrm{move-lower-leg-back}
$$ 
\noindent action often occurs, in relation to a 
$$
\textrm{kick-lower-leg-forward}
$$ 
\noindent action, in the Before (leftmost) slot of an Adjacency ("a") relationship. 

The "no links cross" constraint, in an action grammar context, corresponds essentially to the declaration that a pair of coordinated animations usually relate in an asymmetric way, so that the dependency between two animations often has a direction in which one is "parent" and the other "child", and two animations with different parents do not depend directly on each other.   If one has directional dependency relations, then as Hudson shows in a Word Grammar context, a simple landmark transitivity rule leads to a no-links-crossing constraint (and a more complex landmark transitivity rule, involving landmarks that are not always parents, leads to a slightly looser constraint) \cite{hudson2010}.   

While this requires further study, we have found in the course of our practical robotics work that in the vast majority of everyday movement patterns, there is a directional parent-child relationship between each pair of animations.   For instance, while playing the piano, the "heads" of the animation structure are the hands (for the keyboard) and feet (for the pedals).   The other parts of the body move in ways that are guided by the hand and foot movements.  In effect, then, the overall body movement of piano-playing is parsed into four phrases, for the two hands and two feet.  There are then connections between the heads of these phrases, i.e. the two hands must coordinate with each other, and the hands and feet must coordinate with each other.  But, for instance, the right elbow and left elbow, or the right shoulder and the right knee, do not need to coordinate with each other independently of the hands and feet (if they did, then it would be hard to avoid a "link crossing" situation).

However, it is not clear whether this directional parent-child relationship holds uniformly for all movement patterns that people enact.   Based on a rough analysis it seems to hold generally for yoga poses; e.g. in the tree pose one focuses on one's spine and other body parts are understood to be arranged relative to the spine; in the plank pose the focus is on the core and the feet, which are therefore the "heads" of the animation phrases constituting the overall pose (see \cite{Marion1999} for an effort at formalization of yoga postures, which gives a precise way to represent this sort of idea).    

But modern dance can be quite complex and might contain various exceptions to the no--links-cross constraint, involving tangled relationships instead.   Various formal notations for describing dances exist; some, like Labanotation \cite{HutchinsonGuest1970, Huang2003}, are extremely elaborate and detailed.   Via statistically analyzing multiple dances that have been formalized in Lacanotation, one might infer a formal dance grammar, and could then see if indeed a "no links cross" constraint is obeyed.   It has already been noted that some fairly simple forms of dance such as the foxtrot \cite{hawkins2002} or aerobic exercises \cite{cooper1981} can be modeled using simple grammars; but more complex forms of dance may be a different story.

In general, it seems most sensible to view movement grammars as specifying the overall structure of movements, and to assume that other processes not conveniently specified in grammatical form are going to be used to tune the fine details of movements.   This may be considered analogous to the way the grammar of spoken language determines what words to say, but then other processes figure out the phonological specifics of speech production (and in fact this is more than an analogy, since speech production is also a motor activity carried out via coordination of various muscles in the face).   Regardless of the directions of the dependency relationships in the grammar of animations involved with a movement, the fine-tuning of all the animations involved in a movement is going to involve (explicit or implicit) solution of a system of equations involving parameters of all the different animations.  

So for instance, in the piano-playing example, the role of the hand-animations as the heads of the "arm-animation" phrases, means that for a first approximation, one can figure out what the hands need to do in a certain situation, and then figure out what the elbows and shoulders should do accordingly.   But still this isn't quite how it works, and would lead to awkward movement in some cases.   There is some dependency between what the shoulders do, resulting from the close physical connection of the shoulders.   In fine-tuning the body movements of piano playing, the shoulders must directly coordinate, even though this "crosses links" in terms of the overall animation grammar.   Action grammars are best viewed as providing high-level action plans that constitute guidance for processes of real-time, fine-grained action control.

As with vision or language, the learning of a system of animations and sub-animations suitable for complex real-world control is a complex problem.  For instance, one approach we are exploring in our own research is to co-evolve a reinforcement-learning neural network with an hierarchical neural network, so that the RL network (which is connected to actuators) learns movements to achieve goals, and the hierarchical network models these movements as a hierarchy of animations.   The hierarchical network must be used to infuse the RL network with candidate movement-patterns to try out in attempting to get short-term reward. 

However the learning is done, though, the result is a grammar of movements, with the structure of an asymmetric monoidal category, which can then be mapped morphically into the algebra of logic.   If one articulates a Basic Movement English Link Grammar involving 

\begin{itemize}
\item some nouns and adjectives for use as names of movements
\item words for Allen interval relationships
\item "do" and "done" to connote action
\item the basic connecting words in the Extended RCC/Allen English Link Grammar articulated above
\end{itemize}

\noindent then one can say things such as

{\tt\begin{small}\begin{verbatim}
Kicking the lower leg forward is often done 
after moving the lower leg back.

Pushing the right foot down is done during 
moving the left hand in chord D sharp minor
\end{verbatim}\end{small}}

\noindent  One then can articulate morphic mappings between the movement grammar indicated above, and the set of Allen interval relationships between animation labels, and the sentences of Basic Movement English Link Grammar.   On a formal level this exactly parallels the treatment of image grammar given above.

Morphing action into logic allows action to be morphed into perception, spatiotemporal structure and other structures that can be similarly modeled.   Each of these morphisms captures the key structural information, but omits key quantitative information -- e.g. probabilities in logic, specific geometric patterns of curves and shapes in vision, and specific quantitative parameters of motor movement in action.   But this is as it must be.   Morphisms between different domains allows general knowledge in one domain to propagate to other domains, so that general knowledge becomes in a sense domain independent.  But to turn general knowledge propagated from another domain into specific knowledge, domain-specific quantitative details must then be added.

If one puts together the basic perceptual and action oriented subsets of link grammar identified above, one obtains sentences like

{\tt\begin{small}\begin{verbatim}
After my head is in back of a person, 
kicking the lower leg forward at the person
is often done after moving the lower leg back.

Pushing the right foot down on the pedal,  
is during the pedal being externally connected 
to the right foot in the vertical axis, and during 
moving the left hand in chord D sharp minor
\end{verbatim}\end{small}}

The asymmetric monoidal category corresponding to the grammar of this perception-action language, can be mapped morphically into the closed cartesian category corresponding to the set of logical propositions connecting the animations and visual entity types referred to.   This is a simple exemplification of how logic can be used to glue together perception and action.

Looking at the above example sentences combining perception language and action language, one notes that the syntax gets more complex, and the assumed limitations of the vocabulary cause even more awkwardness than in the simpler examples given before.  It's much nicer, in English, to say "the right foot being on the pedal" rather than "the pedal being externally connected to the right foot in the vertical axis."  But the problem is that, in the former sentence, one then has a tough disambiguation problem; one must map "on" into the RCC relation EC.   

Using the limited vocabulary specified, one has specific words and phrases like "externally connected" and "vertical axis" that map directly into propositional relationships involving spatial and temporal topology.  Natural language achieves greater ease of expression via introducing greater ambiguity, which then introduces greater complexity and uncertainty into the morphisms between language, perception, action and logic.   \footnote{Lojban achieves concise expressions that are suitable for everyday informal human vocal and written communication, and are also  mathematically precise and simplistically mappable into formal structures; but it lacks a sizeable community of speakers.}   But nevertheless, even if one expands consideration to the full grammar of English or another natural language, with all its complexities and ambiguities, the morphisms described here still exist; they are just more lengthy and laborious to articulate in detail.

To highlight the subtlety of natural language ambiguity more explicitly, note that "the right foot being on the pedal" could conceivably refer to a small picture of a right foot being painted on a pedal.  This is an unlikely interpretation but it will be the correct one sometimes.   So one has a situation where a certain syntactic structure maps into more than one logical relational structure.  This does not break the morphic mappings described above, it simply means that the morphism from language into logic is not an isomorphism.  The same linguistic construct may map into more than one different logical construct.   Each mapping has a certain probability associated with it, and there are complex dependencies among the probabilities associated with different interpretations.   Estimating the interdependent probabilities associated with multiple ambiguous linguistic constructs based on tractable heuristic approximations is a subtle and difficult matter, and brings us beyond simple mappings between different domains, and into the realm of generally intelligent language comprehension and usage \cite{Goertzel2014a, Goertzel2014b}, which is well beyond the scope of this paper.

%%%%%%%%%%%%%%%%%%%%%%%%%%%%%%%%%%%%%%%%%%%%%%%%%%%%%%%%%%%%%%%%%%%%%%%%%%%%

\section{Discussion and Future Work}

The construction of practical systems for processing linguistic and spatiotemporal data, and grounding the former in the latter, involves numerous complexities due to the need to maximize pattern recognition accuracy within the scope of limited computing resources.   However, the abstract structure of the mappings from linguistic and spatiotemporal structure to and from logic take a simple and elegant form, which we have articulated here for the first time, building on earlier practical and theoretical work.   Similar mappings appear to exist in the context of other cognitive domains such as action and vision, and many others. 

\begin{figure}[H]
\begin{center}
\includegraphics[width=14cm]{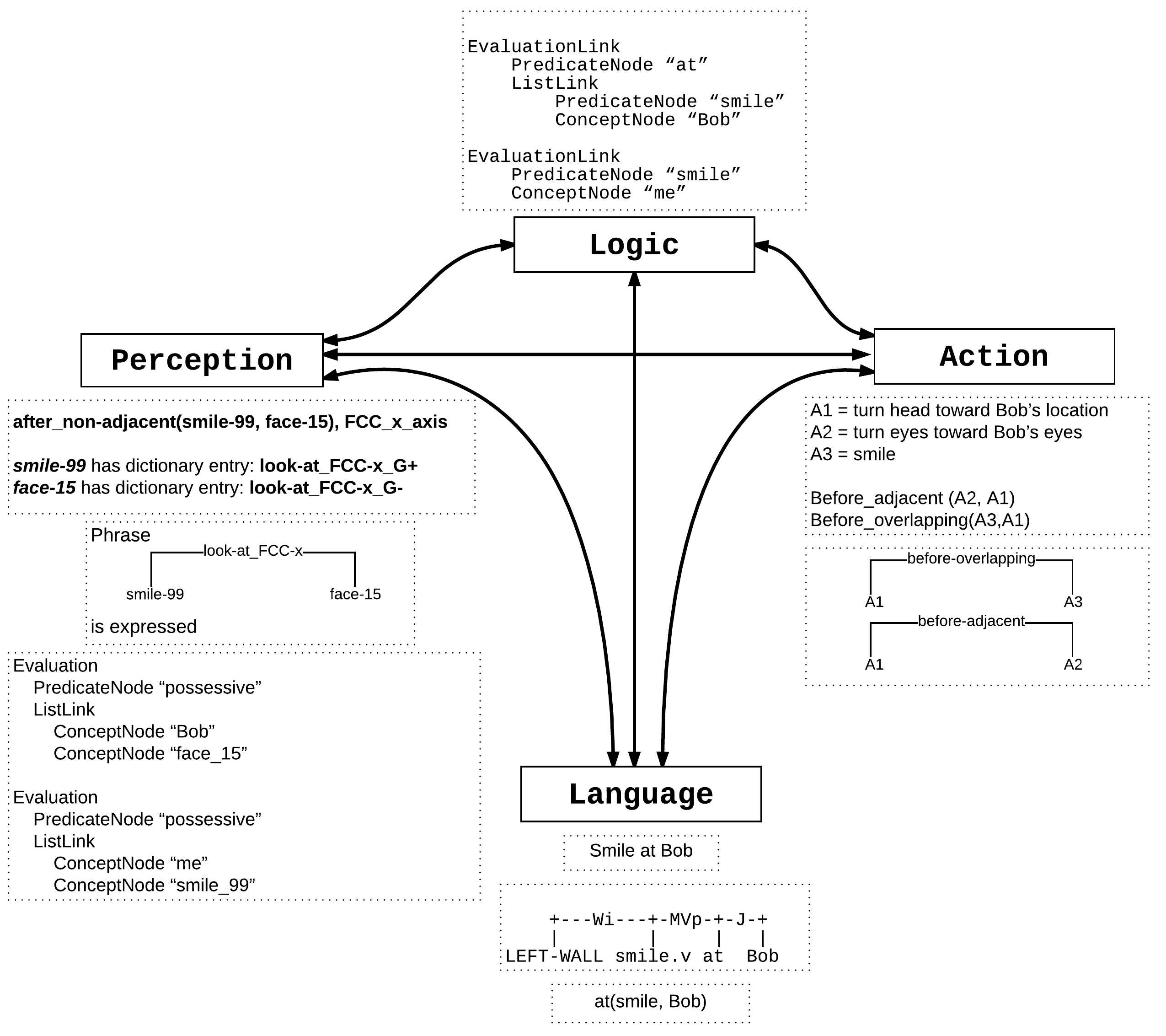}
\caption{Depiction of the  relationship "smile at Bob" as represented in the languages of: logic, perception, action and natural language.   In each case the representation involves a combination of two  parts.   The arrows indicate morphisms between the different domains; and each of these morphisms preserves the relationship between the two parts involved in the relationship.}
\label{fig:newDiagram}
\end{center}
\end{figure}

Figure \ref{fig:newDiagram} provides a simple illustration of these ideas, showing how a simple relationship ("Smile at Bob") manifests itself as a combination of two sub-relationships, in the four domains of perception, action, logic and natural language.   Morphisms between the different domains preserve the logic of the decomposition into sub-relationships.   Among other purposes, these morphisms ground the linguistic construct "smile at Bob" in relevant, morphically structured perceptions and actions; and abstract the linguistic construct into a pair of logical representations denoting (key aspects of) its meaning.

In the context of AI system design, the value of constructing systems whose knowledge representations embody these abstract mappings, is that this can ease the design and implementation of learning and reasoning methods that span and bridge the domains of language, spacetime and logic in a natural way, without requiring a great amount of explicit bridging code.   As an example, in the OpenCog AI system, if we use Probabilistic Logic Networks (PLN) inference to generalize from observed examples in either the linguistic or the spatiotemporal domain, the mappings between language and spacetime via logic can be used to propagate the uncertain generalizations from one domain to another.   Concrete elucidations of this propagation will be described in follow-up publications.

%%%%%%%%%%%%%%%%%%%%%%%%%%%%%%%%%%%%%%%%%%%%%%%%%%%%%%%%%%%%%%%%%%%%%%%%%%%%
%\section*{References}

%\begin{thebibliography}

\bibliographystyle{alpha}
\bibliography{ref-1}

%\end{thebibliography}

\end{document}